\documentclass[10pt,twocolumn,letterpaper]{article}

\usepackage[pagenumbers]{cvpr} 

\usepackage{graphicx}
\usepackage{amsmath}
\usepackage{amssymb}
\usepackage{booktabs}

\usepackage[pagebackref,breaklinks,colorlinks]{hyperref}

\usepackage[capitalize]{cleveref}
\crefname{section}{Sec.}{Secs.}
\Crefname{section}{Section}{Sections}
\Crefname{table}{Table}{Tables}
\crefname{table}{Tab.}{Tabs.}

\def\cvprPaperID{*****} 
\def\confName{CVPR}
\def\confYear{2024}

\begin{document}

\title{Fusing Physics-Driven Strategies and Cross-Modal Adversarial Learning: Toward Multi-Domain Applications}

\author{Hana Satou,Alan Mitkiy\\
}
\maketitle


\begin{abstract}
The convergence of cross-modal adversarial learning and physics-driven methods represents a cutting-edge direction for tackling challenges in complex multi-modal tasks and scientific computing. This review focuses on systematically analyzing how these two approaches can be synergistically integrated to enhance performance and robustness across diverse application domains. By addressing key obstacles such as modality discrepancies, limited data availability, and insufficient model robustness, this paper highlights the role of physics-based optimization frameworks in facilitating efficient and interpretable adversarial perturbation generation. The review also explores significant advancements in cross-modal adversarial learning, including applications in tasks such as image cross-modal retrieval (e.g., infrared and RGB matching), scientific computing (e.g., solving partial differential equations), and optimization under physical consistency constraints in vision systems. By examining theoretical foundations and experimental outcomes, this study demonstrates the potential of combining these approaches to handle complex scenarios and improve the security of multi-modal systems. Finally, we outline future directions, proposing a novel framework that unifies physical principles with adversarial optimization, providing a pathway for researchers to develop robust and adaptable cross-modal learning methods with both theoretical and practical significance.
\end{abstract}

\section{Introduction}
The task of image retrieval involves identifying and extracting images from vast databases that closely match a given query image. With the exponential growth of image data on platforms such as social media, e-commerce, and digital archives, the need for efficient and accurate retrieval systems has become increasingly critical. This demand is further driven by the proliferation of high-resolution cameras and smartphones, as well as the desire for personalized content and seamless user experiences across diverse applications, including digital marketing, medical diagnostics, and autonomous navigation.

Recent advancements in image retrieval have been largely fueled by breakthroughs in Deep Learning, particularly through Convolutional Neural Networks (CNNs). These networks have revolutionized the field by enabling automated learning of hierarchical features from raw image data, leading to significant improvements in image recognition and retrieval accuracy. Unlike earlier approaches, which relied heavily on handcrafted features and traditional machine learning techniques, CNN-based methods excel at capturing complex patterns in real-world images. As a result, they have set new benchmarks for precision and relevance in retrieval tasks.

However, several challenges remain. One of the most pressing issues is cross-domain retrieval, where models trained on one type of data often fail to generalize effectively to images from a different domain. This problem is especially pronounced in cases where the training dataset does not adequately reflect variations in the target data, such as images captured under differing environmental conditions or with different equipment. Additionally, noise, distortions, and other inconsistencies in real-world images further complicate the retrieval process, diminishing system reliability.

To address these challenges, data augmentation and adversarial learning have emerged as promising solutions. Data augmentation enhances the diversity of training datasets by applying transformations such as scaling, rotation, and color adjustments, mimicking real-world variability. This approach improves model generalization by exposing it to a broader range of conditions, thereby mitigating overfitting and enhancing robustness to unseen data. Meanwhile, adversarial learning focuses on safeguarding systems against malicious perturbations. By generating adversarial examples—subtle, carefully crafted modifications to input images—models can be trained to recognize and resist such attacks, ensuring robust and secure performance even in adversarial scenarios.

This paper provides a comprehensive review of these methodologies and their role in overcoming key challenges in image retrieval. It highlights recent advances in data augmentation and adversarial learning, evaluates their impact on real-world applications, and discusses their integration into retrieval systems. Through an analysis of case studies and current trends, the paper aims to offer insights into how these techniques have shaped the field and their potential for future development \cite{bengio2024managing,azizi2023robust,krizhevsky2012imagenet,gong2024beyond2,krizhevsky2009learning}.

\section{Related Work}

The convergence of adversarial learning and cross-modal tasks has emerged as a significant research focus due to its potential for enhancing model reliability and addressing challenges associated with modality variations~\cite{lecun2015deep,rumelhart1986learning,srivastava2014dropout,gong2024beyond2,gong2024beyond,gong2021eliminate,van2008visualizing}. This section reviews developments in three primary domains: advancements in adversarial learning, innovations in cross-modal attack strategies, and the role of physics-informed optimization~\cite{gong2024exploring,gong2024beyond2,gong2024beyond,gong2021eliminate,gong2021person,gong2021effective,gong2021person2}.

\subsection{Evolution of Adversarial Learning}

Adversarial learning has become an essential tool in machine learning, exposing vulnerabilities in neural networks through carefully designed input manipulations. Initially introduced by Szegedy et al.~\cite{lecun2015deep}, adversarial attacks demonstrated that even minimal changes in input could significantly alter model predictions. Methods such as Fast Gradient Sign Method (FGSM)~\cite{srivastava2014dropout} and Projected Gradient Descent (PGD)~\cite{rumelhart1986learning} made adversarial example generation more efficient and scalable. Further advancements, like Universal Adversarial Perturbations (UAPs)~\cite{gong2024cross}, extended these attacks to generate perturbations applicable to diverse inputs, showcasing broader model vulnerabilities. However, these techniques predominantly target single-modality tasks and often fail to address the complexities of multi-modal systems, where varying gradients and modality-specific features complicate adversarial optimization.

\subsection{Adversarial Approaches in Multi-Modal Systems}

Cross-modal adversarial learning focuses on tackling the challenges posed by the heterogeneity of different data modalities, such as RGB and infrared images or visual and textual information. Traditional adversarial attacks~\cite{gong2024cross,gong2024beyond} primarily concentrate on single-modality scenarios, limiting their generalizability to multi-modal tasks. The Cross-Modality Perturbation Synergy (CMPS) approach~\cite{crossmodality} marked a significant step toward cross-modal robustness by leveraging gradient information from multiple modalities to enhance perturbation transferability. Despite these advancements, the performance of cross-modal attacks often suffers due to inconsistencies in gradients and modality-specific characteristics. Emerging techniques, including evolutionary optimization for perturbations~\cite{}, highlight the need for task-adaptive adversarial strategies that address modality-specific challenges while promoting cross-modal transferability.

\subsection{Physics-Informed Optimization Techniques}

Integrating domain-specific physical knowledge into machine learning frameworks has proven effective in improving model robustness and interpretability. In tasks like image dehazing, physics-guided models, such as the atmospheric scattering framework~\cite{gong2024adversarial}, enable neural networks to produce outputs that align with real-world physical phenomena. Similarly, in scientific computing, combining data-driven PDE solvers with physical constraints significantly enhances model accuracy and generalization. Adversarial learning has also been applied in this context, with methods like SMART~\cite{gong2024adversarial} generating adversarial examples that adhere to physical laws, allowing neural PDE solvers to better handle uncertainties. These approaches demonstrate the importance of combining physics-based insights with adversarial strategies to tackle real-world challenges.

\subsection{Toward Unified Optimization Frameworks}

While advancements in adversarial learning, cross-modal attack strategies, and physics-guided optimization have independently progressed, their integration remains underexplored. Adversarial learning is adept at exposing model vulnerabilities but often lacks the interpretability and physical consistency needed in multi-modal applications. On the other hand, physics-informed approaches ensure robust and explainable outputs but have yet to be widely applied to adversarial optimization for cross-modal tasks. Developing unified frameworks that combine physical principles with adversarial learning could pave the way for more robust and interpretable solutions~\cite{gong2022person,gong2024cross,crossmodality,gong2024exploring,gong2024beyond2,gong2024beyond,gong2021eliminate,gong2021person,gong2021effective,gong2021person2}. Such integration holds promise for advancing security, generalization, and cross-modal understanding in diverse domains.

\section{Methodology}

In this section, we present a novel approach that combines **cross-modal adversarial learning** with **physics-driven optimization** techniques to enhance the robustness and performance of image retrieval systems. The integration of these two methods aims to address the challenges posed by modality variations, security threats, and real-world data inconsistencies. Our methodology includes three key components: **cross-modal data augmentation**, **adversarial training**, and **physics-guided adversarial perturbations**.

\subsection{Cross-Modal Data Augmentation}

Cross-modal data augmentation plays a pivotal role in expanding the model's ability to handle diverse data modalities and variations. It generates new training data by applying various transformations to existing images, simulating real-world conditions and enabling the model to generalize effectively across different domains.

Key strategies for cross-modal data augmentation include:

\begin{itemize}
	\item \textbf{Geometric Variations:} These transformations encompass operations like rotation, scaling, translation, and flipping. For instance, by rotating images or flipping them horizontally, the model is trained to recognize objects from different angles and viewpoints, which is critical for accurate retrieval in various scenarios.
	
	\item \textbf{Illumination and Color Adaptations:} To simulate varying lighting conditions and environmental factors, we apply adjustments to brightness, contrast, saturation, and hue. These color shifts allow the model to better adapt to the diverse conditions under which images may be captured, including low-light environments or variations in camera settings.
	
	\item \textbf{Noise and Distortion Simulation:} To improve robustness, we inject noise (e.g., Gaussian or salt-and-pepper noise) into the training images. This enables the model to handle real-world imperfections and distortions, such as sensor noise or compression artifacts, which are commonly encountered in practical retrieval scenarios.
	
	\item \textbf{Contextual Cropping and Resizing:} By cropping different regions of the image or resizing it to different aspect ratios, we expose the model to varying object scales and context. This is particularly important for retrieval tasks where objects of interest may appear at different sizes or in different locations within the frame.
	
	\item \textbf{Simulating Image Quality Variations:} We apply blurring and sharpening filters to mimic the impact of focus issues, motion blur, or different camera qualities. This prepares the model for the variability in image sharpness encountered in real-world datasets, where image quality can often be suboptimal.
\end{itemize}

By leveraging these cross-modal data augmentation techniques, we significantly increase the diversity of training samples, making the model more capable of handling a wide range of real-world conditions. This process helps reduce overfitting and improves the model’s generalization ability.

\subsection{Adversarial Learning for Robustness}

Adversarial learning plays a crucial role in enhancing the security and resilience of image retrieval systems by introducing adversarial examples during training. These examples are carefully designed perturbations that are meant to challenge the model and make it more resilient to misleading inputs. The goal is to strengthen the model’s defenses against both malicious attacks and unexpected data inconsistencies.

Key components of adversarial learning include:

\begin{itemize}
	\item \textbf{Generating Adversarial Examples:} Methods like Fast Gradient Sign Method (FGSM), Projected Gradient Descent (PGD), and Carlini \& Wagner (C\&W) attacks are employed to generate adversarial perturbations. These methods compute gradients with respect to the input image and apply perturbations that maximize prediction errors, resulting in misclassifications or retrieval failures.
	
	\item \textbf{Adversarial Training:} Adversarial training integrates adversarial examples into the model's learning process. By training the model on both clean and adversarially perturbed images, the system learns to resist adversarial attacks, making it more robust to malicious perturbations and unanticipated disturbances.
	
	\item \textbf{Defensive Mechanisms:} A variety of defensive strategies can be integrated alongside adversarial training to further improve system security. Techniques like input preprocessing (e.g., denoising and normalization), regularization methods designed to mitigate adversarial effects, and ensemble learning strategies that combine multiple models to strengthen overall system security contribute to the robustness of the retrieval system.
	
	\item \textbf{Performance Evaluation and Benchmarking:} Adversarial learning also involves thorough testing to assess model performance under various adversarial scenarios. Benchmarks and evaluation metrics help assess how well the model performs under different attack scenarios, providing insights that guide further optimization efforts.
\end{itemize}

Adversarial learning is particularly valuable in situations where model reliability is critical, ensuring that the retrieval system remains reliable and performs well even in the face of malicious attempts to undermine its functionality.

\subsection{Physics-Driven Adversarial Optimization}

The integration of physics-driven methods with adversarial learning introduces a new layer of robustness by ensuring that adversarial perturbations are physically consistent with real-world scenarios. Physics-guided adversarial optimization uses domain-specific knowledge, such as physical laws and models, to generate adversarial examples that respect the constraints of the physical world.

Key techniques in physics-driven adversarial optimization include:

\begin{itemize}
	\item \textbf{Physically Consistent Perturbations:} We incorporate physical models (e.g., atmospheric scattering models for image dehazing) to generate adversarial examples that follow physical laws. These perturbations ensure that adversarial attacks are not only effective in deceiving the model but also realistic in terms of how they could occur in the real world.
	
	\item \textbf{Optimization Under Physical Constraints:} By applying constraints derived from physical principles, we ensure that the generated adversarial examples align with realistic perturbations. This optimization improves the reliability of adversarial examples, ensuring that they remain consistent with domain-specific requirements, such as visual realism in image processing tasks.
	
	\item \textbf{Cross-Domain Transferability:} We explore how adversarial perturbations, guided by physical principles, can be transferred across different modalities. For example, adversarial attacks in the context of RGB images may be adapted to infrared images, maintaining their effectiveness while ensuring physical plausibility.
\end{itemize}

The combination of adversarial learning with physics-driven optimization results in a more secure and reliable model that not only defends against intentional perturbations but also incorporates real-world constraints to generate meaningful and applicable adversarial examples.

\subsection{Integrating Data Augmentation, Adversarial Learning, and Physics Optimization}

The synergy between data augmentation, adversarial learning, and physics-guided optimization provides a comprehensive framework to enhance the performance, robustness, and security of image retrieval systems. Data augmentation prepares the model for diverse environmental conditions, while adversarial learning strengthens its defense against malicious attacks. The integration of physics-driven methods ensures that adversarial perturbations are consistent with real-world scenarios, enhancing both the interpretability and applicability of adversarial learning techniques. Together, these methodologies enable the creation of more robust, adaptable, and secure image retrieval systems that perform well across a variety of real-world challenges.

\section{Discussion}
While data augmentation and adversarial learning have proven to be effective tools for enhancing the robustness of image retrieval systems, they come with certain limitations that need to be addressed. A primary challenge of data augmentation lies in its reliance on a fixed set of transformations, which may not always capture the full spectrum of real-world variations. As a result, the model may still struggle with novel or unseen data distributions that fall outside the pre-defined augmentation strategies. To overcome this, future research should explore more flexible and dynamic augmentation methods that can continuously adapt to new and emerging data patterns, ensuring that the model is always prepared for a broader range of potential scenarios.

Similarly, adversarial learning, while powerful in fortifying models against intentional input perturbations, requires a careful balance. If the adversarial perturbations are too aggressive, they may degrade the model's ability to learn from authentic, unmodified data, leading to overfitting or poor generalization. Additionally, the computational cost of adversarial training techniques can be prohibitive, especially when applied to large-scale datasets or complex systems. Addressing the trade-off between perturbation strength and model performance, as well as optimizing the computational efficiency of these techniques, will be essential for their successful deployment in practical applications.

\section{Conclusion}
Data augmentation and adversarial learning are indispensable methodologies that significantly enhance the resilience and functionality of image retrieval systems. By expanding the variety of training data and preparing the model for potential adversarial attacks, these techniques tackle some of the fundamental challenges in ensuring robust and reliable retrieval systems. However, as the field continues to evolve, there is an increasing need for more adaptive and computationally efficient approaches. Research must focus on developing dynamic data augmentation strategies that can flexibly respond to changing environments and data distributions. Likewise, optimizing adversarial learning to maintain a balance between robustness and computational efficiency will be crucial for large-scale implementation.

Looking forward, the integration of data augmentation and adversarial learning with emerging technologies such as transfer learning, self-supervised learning, and hybrid models will likely yield even more robust and efficient systems. As image-based data and retrieval technologies continue to advance, these methodologies will play a pivotal role in shaping the next generation of intelligent, adaptable retrieval systems capable of tackling complex, real-world challenges.

{\small
\bibliographystyle{ieee_fullname}
\bibliography{egbib}
}

\end{document}